\documentclass[sigconf]{acmart}
\AtBeginDocument{%
  \providecommand\BibTeX{{%
    \normalfont B\kern-0.5em{\scshape i\kern-0.25em b}\kern-0.8em\TeX}}}

\usepackage{algorithm}
\usepackage{algpseudocode}
\usepackage{bm}
\usepackage{caption}
\usepackage{subcaption}
\usepackage{graphics}
\usepackage{graphicx}
\usepackage{adjustbox}
\begin{document}

\title{DeeprETA: An ETA Post-processing System at Scale}
\author{
  Xinyu Hu, Tanmay Binaykiya, Eric Frank, Olcay Cirit
}
\email{{xinyuh,btanmay,mysterefrank, olcay}@uber.com}
\affiliation{%
  \institution{Uber Technologies Inc}
  \city{San Francisco}
  \state{CA}
  \country{USA}
}

\renewcommand{\shortauthors}{Hu, Binaykiya, Frank and Cirit}

\begin{abstract}
Estimated Time of Arrival (ETA) plays an important role in delivery and ride-hailing platforms. For example, Uber uses ETAs to calculate fares, estimate pickup times, match riders to drivers, plan deliveries, and more. Commonly used route planning algorithms predict an ETA conditioned on the best available route, but such ETA estimates can be unreliable when the actual route taken is not known in advance. In this paper, we describe an ETA post-processing system in which a deep residual ETA network (DeeprETA) refines naive ETAs produced by a route planning algorithm. Offline experiments and online tests demonstrate that post-processing by DeeprETA significantly improves upon the accuracy of naive ETAs as measured by mean and median absolute error. We further show that post-processing by DeeprETA attains lower error than competitive baseline regression models. 

\end{abstract}





\maketitle

\section{Introduction}
Each year, ride-hailing and delivery platforms such as Uber, Didi, Doordash and others power billions of transactions that depend critically on accurate arrival time predictions (ETA). For example, Uber uses ETAs to calculate fares, estimate pickup and dropoff times, match riders to drivers and couriers to restaurants.  Due to the sheer volume of decisions informed by ETAs, reducing ETA error by even low single digit percentages unlocks tens of millions of dollars in value per year by increasing marketplace efficiency.

A common approach to predict ETAs is with a routing engine, which is also called a route planner. Routing engines divide up the road network into small road segments represented by weighted edges in a graph. They use shortest-path algorithms\cite{Dijkstra1959} to find the best path from origin to destination and add up the weights to obtain an ETA. 
Modern routing engines even take into account real-time traffic patterns, accidents and weather when estimating the time to traverse each road segment. Still, graph-based models used by routing engines can be incomplete with respect to real-world planning scenarios typically encountered in ride-hailing and delivery: 
\begin{itemize}
    \item \textbf{Route uncertainty}. We generally don't know in advance which route a driver or courier will choose to take to their destination, yet we still require an accurate ETA that factors this uncertainty in account.
    \item \textbf{Human error}. Human drivers make mistakes especially in difficult sections of the road network, but shortest-path algorithms can't account for this. 
    \item \textbf{Distribution Shift}. Empirical arrival time distributions differ markedly across different tasks, such as driving to a restaurant or driving to pick up a rider, even when the shortest path is the same. 
    \item \textbf{Uncertainty estimation}. Different ETA use cases call for distinct point estimates of the predictive distribution. For example, fare estimation requires a mean ETA, whereas user-facing ETAs may call for a set of ETA quantiles or expectiles.
\end{itemize} 
 
 The notable shortcomings of routing engine ETAs create an opportunity to use observational data to produce ETAs better aligned with business needs and real-world outcomes. Past ML-based approaches to ETA prediction for ride hailing and food delivery assume that the route will be fixed in advance \cite{fu2020compact}, or they combine route recommendation with ETA prediction \cite{gao2021adeep}. Both past approaches solve simplified versions of the ETA prediction problem by making assumptions about the route taken or the type of task.

\begin{figure}[h]
  \centering
  \includegraphics[width=\linewidth]{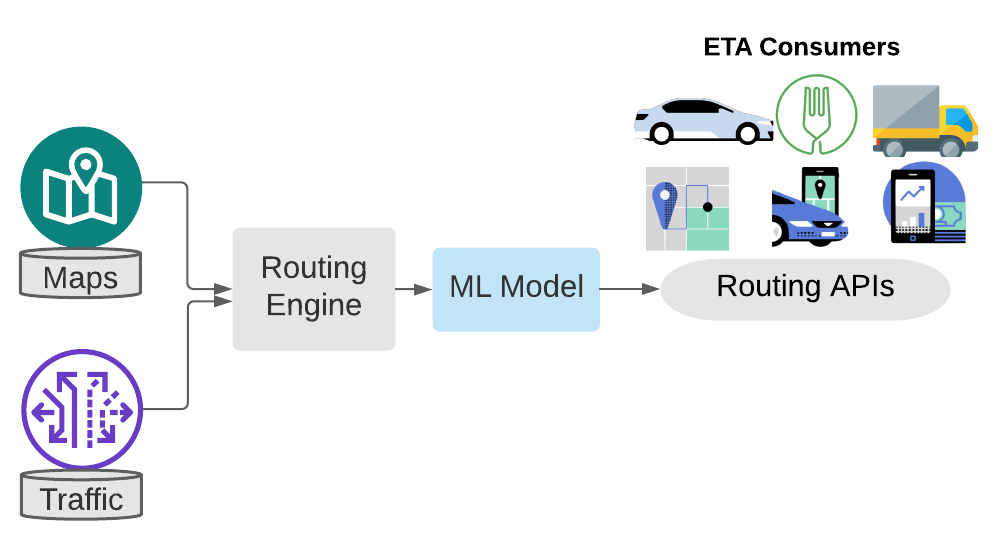}
  \caption{Hybrid approach of ETA post-processing using ML models }
  \label{fig:key_components}
\end{figure}

We take a hybrid approach, termed ETA post-processing, that treats the routing engine ETA as a noisy estimate of the true arrival time. We use a deep residual ETA network, referred to as DeeprETANet, to predict the difference between the routing engine ETA and the observed arrival time. The key components of our solution, shown in Figure \ref{fig:key_components}, include a routing engine with access to maps and real-time traffic, and a machine learning model that refines the routing engine ETAs. We show that this hybrid approach outperforms both the unaided routing engine as well as competitive baseline regression models. Our proposed approach can be implemented on top of any routing engine and is intended for low-latency, high-throughput deployment scenarios. 

Compared to other works that predict estimated arrival times, the key contributions of this paper are 
\begin{itemize}
    \item \textbf{ETA Post-processing}. Our problem formulation, described in section \ref{section:problemformulation}, which treats programmatic ETAs from a route planner as noisy estimates of true arrival times, is unique in the travel-time estimation literature.
    \item \textbf{DeeprETANet Architecture}. A deep learning architecture for ETA post-processing, described in section \ref{section:PP}, that improves ETA accuracy compared to strong regression baselines while adding minimal incremental serving latency.
    \item \textbf{Multi-resolution Geospatial Embeddings}. A scheme for embedding geospatial location information using multiple independent hash functions for each spatial resolution, in section \ref{section:embeddingmodule}. 
\end{itemize}

\section{Related Work}

In this section we describe works that influenced our design or are part of the broader field of travel time estimation with deep learning.

\textit{Travel Time Estimation}: 
 Regression-based approaches predict an ETA from a set of features describing a trip, such the origin and destination, or segment-level route information. Past works of this kind have used combinations of convolutional and recurrent architectures  \cite{wang2018learning, wang2018deeptte}, ResNets \cite{li2018murat}, graph neural networks \cite{hong2020heteta, derrow-pinion2021eta, fu2020compact} and self-attention \cite{sun2021fmaeta}. Other works go beyond travel time estimation by jointly learning to predict routes   \cite{zheng2021solving, gao2021adeep}. DeeprETA is a hybrid regression model that learns a residual correction on top of predictions made by a routing engine. Because it is optimized for speed, it uses a shallow encoder-decoder architecture with a linear transformer \cite{katharopoulos2020transformers} to encode features. 

\textit{Geospatial Embeddings:} Arrival times depend heavily on the origin and destination locations. Past works have encoded these locations with multi-scale sinusoidal embeddings \cite{mai2020multiscale} or have used LSTMs to learn geospatial embeddings \cite{wu2019deepeta}. \cite{zheng2021solving} use grid embeddings of latitude and longitude in which the grid cell is augmented with the relative distances between the original point and the four corners of grid. In \cite{li2018murat}, latitude and longitude are embedded separately over uniform grids to reduce cardinality, and graph pre-training is used to incorporate road network connectivity information for each origin and destination.  DeeprETA uses feature hashing \cite{weinberger2009feature, moody1988fast} to reduce geospatial cardinality:  similar to \cite{moody1988fast}, we discretize location features into multiple grids with varying resolutions, but uniquely apply multiple independent hash functions to each grid cell and learn an embedding in each hash bin rather than just a weight. As in \cite{saberian2019gradient} we use quantile binning to pre-process continuous features into discrete bins, although we learn embeddings over each bin in lieu of one-hot encoding.  

\textit{Feature encoding:} A wide and deep recurrent network was proposed to capture spatial-temporal information \cite{wang2018learning}. This model uses a cross-product of embeddings to learn feature interactions, which is commonly used in recommendation systems \cite{cheng2016wide, guo2017deepfm, zhou2018deep}. These works motivate us to design a model with smart ways of learning feature interaction.

\section{Problem formulation}
\label{section:problemformulation}
We define the ETA prediction task as predicting the travel time from a point A to a point B. The travel path between A to B is not fixed as it may depend on the real-time traffic condition and what routes drivers or couriers choose. Therefore, we propose a hybrid post-processing approach for ETA predictions, which consists of two main components: a routing engine and a ML post-processing module. 

We formulate ETA prediction as a regression problem. The label is the actual arrival time (ATA), which is a continuous positive variable, denoted as $Y \in R^{+}$. The ATA definition varies by the request type, which could be a pick-up or a drop-off. For a pick-up request, the ATA is the time taken from driver accepting the request to beginning the ride or delivery trip. For a drop-off request, the ATA is measured from the begin of the trip to the end of the trip.

The ETA from the routing engine is referred to as the routing engine ETA (RE-ETA), denoted as $Y_0 \in R^{+}$. For each ETA request $\bm{q}_i = \{\tau_i, \bm{p}_i, \bm{x}_i\}$, where $\tau_i$ is the timestamp, $\bm{p}_i = \{p_{i1}, p_{i2}, \cdots, p_{im}\}$ is the recommended route segments from the routing engine and $\bm{x}_i$ is the features at timestamp $\tau_i$, for $i=1,\dots, n$. Our task is to learn a function $\mathcal{G}$ that maps the features to predict an ATA. The ETA is denoted as $\hat{Y}$:
\begin{equation}
 \mathcal{G}(\bm{q}_i) \to \hat{y}_i. 
  \label{eqn:eta}
\end{equation}

The recommended route from the routing engine is represented as a sequence of road segments $\bm{p}_i$. The RE-ETA is calculated by summing up traversal time of the road segments $\hat{y}_{0i} = \sum_{j=1}^m t_{p_{ij}}$,  where $t_{p_{ij}}$ denotes the traversal time of the $j$th road segment $p_{ij}$ for the $i$th ETA request. The quality of $\hat{y}_{0i}$ depends on the estimated road segments $t_{p_{ij}}$. In real world situations, drivers may not always follow the recommended route, resulting in re-routes. Therefore $\hat{y}_{0i}$ may not be an accurate estimation of ATA. 

\begin{figure}[h]
  \centering
  \includegraphics[width=\linewidth]{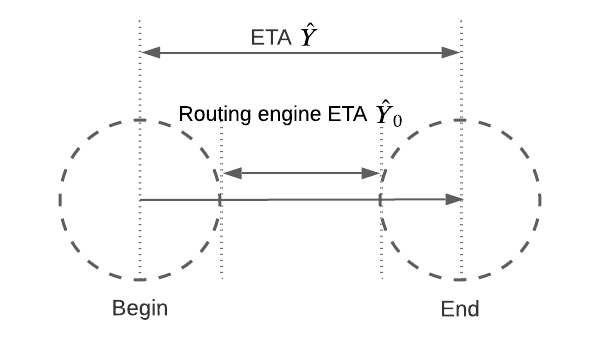}
  \caption{ETA vs routing engine  ETA}
    \label{fig:unmodified_eta}
\end{figure}

The begin and end location neighborhoods of a request account for a large proportion of noise. For instance, the driver may have to spend time in looking for a parking spot. The RE-ETA circumvents this uncertainty by estimating the travel time between the neighborhoods instead of the actual begin and end locations. Figure ~\ref{fig:unmodified_eta} illustrates the difference between RE-ETA and the final ETA. The focus of this paper is not on improving the route recommendation. Therefore, we will not cover the details of estimating the RE-ETA $\hat{y}_0$. To accommodate the difference between ETA $\hat{y}$ and RE-ETA $\hat{y}_0$, we need a post-processing system to process the $\hat{y}_0$ for more accurate ETA predictions,
\begin{equation}
 \mathcal{G}(\bm{q}_i, \hat{y}_0) \to \hat{y}_i. 
  \label{eqn:posteta}
\end{equation}

Accurate ETA depends on the quality of map data and real-time traffic signals. For an ETA request $q_i$, $X_i \in R^p$ denotes a $p$-dimensional feature vector. $X_i$ includes the features listed in Table \ref{tab:feature}. Among all, the spatial and temporal features are the most important ones. These include origin, destination, time of the request, as well as information about real-time traffic and the nature of the request, i.e food delivery drop-off vs ride-hailing pick-up.

\begin{table}
  \caption{Raw features in one ping}
  \label{tab:feature}
  \begin{tabular}{cl}
    \toprule
    Raw Features & Description\\
    \midrule
    Temporal Feature &  minute of day, day of week, etc. \\
        \midrule
    Geospatial Feature & latitude and longitude of begin/end location\\
        \midrule
    Trip & trip type, route type, request type\\
     & routing engine  ETA \\
     & estimated distances \\
         \midrule
    Traffic Feature & real-time and historical speed\\
        \midrule
    Context & country/region/city IDs, etc.\\
    \midrule
    Label & actual time of arrival\\
  \bottomrule
\end{tabular}
\end{table}

\section{DeeprETA post-processing system}
\label{section:PP}
We propose a post-processing system named DeeprETA to tackle the ETA prediction task. DeeprETA post-processing system aims at predicting the ATA by estimating a residual that added on top of the routing engine ETA. This post-processing step is separated from the route recommendation from the routing engine. The residual $\hat{r}_i$ is a function of $(\bm{q}_i, \hat{y}_{0i})$ to correct the RE-ETA, 
\begin{equation}
 \hat{y}_i =  \hat{y}_{0i} + \hat{r}_i. 
  \label{eqn:eta_residual}
\end{equation}

\begin{figure}[h]
  \centering
  \includegraphics[width=\linewidth]{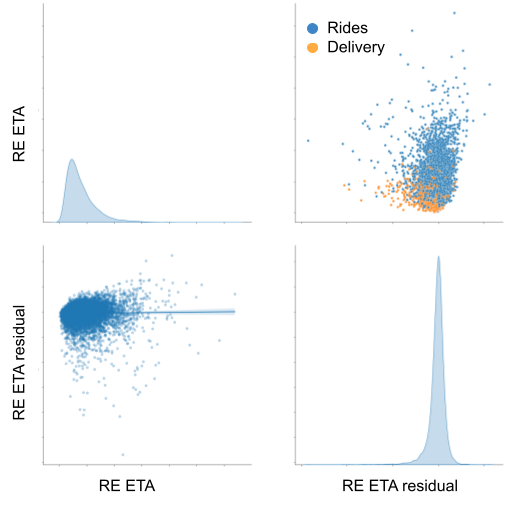}
  \caption{Distribution of RE-ETA and its residuals on sampled data. Top left: distribution of RE-ETA; Top right: scatter plot of RE-ETA vs RE-ETA residuals for rides and eats requests; Bottom left: regression plot of RE-ETA vs RE-ETA residuals; Bottom right: distribution of RE-ETA residuals.}
  \label{fig:reeta_residual}
\end{figure}

It is challenging to design this post-processing system due to the following reasons: firstly, the RE-ETA data and the residual distribution are skewed and have long tails. This is shown in the top left and bottom right in Figure \ref{fig:reeta_residual}. Log-transformation is usually used for normalizing skewed data distribution. However transforming the log-scale prediction back to the original scale by exponentiation usually results in some extreme values, which may affect user experience. We propose to solve this issue by using an asymmetric loss function. The details are covered in section \ref{section:train}. Secondly, heterogeneity of data is due to multiple trip/request types; for example, the RE-ETA distributions of a rides trip and a delivery trip are different, as illustrated in top right in Figure \ref{fig:reeta_residual}. Treating a ride and a delivery trip the same will result in a less accurate regression model, as shown in bottom left of Figure \ref{fig:reeta_residual}. Therefore, we design a model structure to deal with trip heterogeneity specifically. Thirdly, this post-processing system needs to handle a large volume of requests with a low latency. Ride-hailing platforms, like Uber, has billion-level requests to process on a weekly basis. To be able to serve real-time online, the system has a stringent requirement on latency.

In this section, we first describe the overall post-processing architecture, then we introduce the DeeprETANet architecture with strategies we adopt for improve the accuracy and latency. Finally, we describe the training and serving details of DeeprETA.

\subsection{DeeprETANet Overview}
\label{subsec:DeeprETANet}
\begin{figure}[h]
  \centering
  \includegraphics[width=\linewidth]{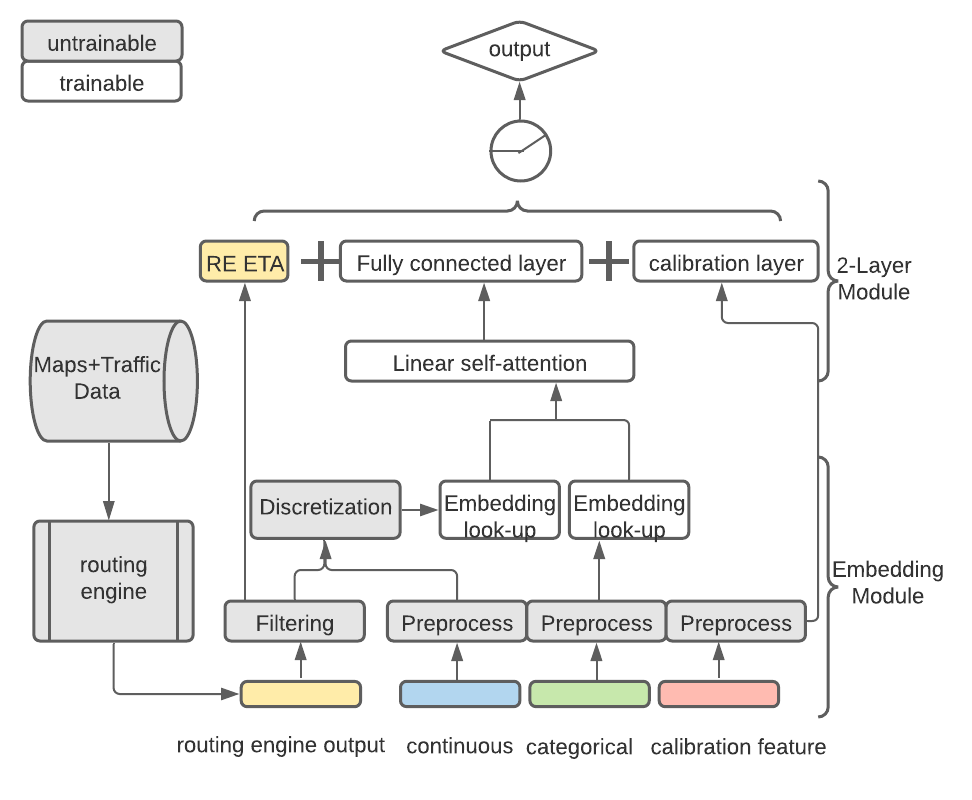}
  \caption{DeeprETA Post-processing Architecture}
  \label{fig:architecture}
\end{figure}

We refer to the model used for post-processing as DeeprETANet. DeeprETANet consists of two modules: an Embedding module and a Two-layer Module. The high-level architecture is illustrated in Figure \ref{fig:architecture}. The embedding module takes in outputs from the routing engine and features associated with the ETA requests. We categorize the features into three categories: continuous features, categorical features and calibration features as shown in Figure \ref{fig:architecture}. The calibration features convey different segments of the trip population such as whether it is a delivery drop-off or ride-sharing pick-up trip. All features are pre-processed into discrete bins prior to the embedding module that learns the hidden representation of each bin. The interaction module aims to learn the interaction of spatial and temporal embeddings. Next a decoder with a calibration layer is used to calibrate the predicted residuals based on the request type. Finally the calibrated residual is added to the RE-ETA as the predicted ETA. 

\subsection{Embedding Module}
\label{section:embeddingmodule}
The embedding is the cornerstone of DeeprETANet for the following two reasons: accuracy and speed. In our experiments, we found discretizing and embedding all the features provided a significant accuracy lift over using continuous features directly. For the speed, although these embeddings make up 99\% of the hundreds of millions of parameters in DeeprETA, they contribute minimally to latency, because embedding look-up table takes $O(1)$ time. 

We map all categorical and continuous features to embeddings. Suppose we have a dataset with $n$ instances, each has a feature vector $\bm{x}_i$. It contains $p$ dimensional features $\bm{x}_i = [x_{i1}, x_{i2}, \cdots, x_{ip}]$.

\textbf{Categorical features}: The embedding of a categorical feature $\bm{x}_{\alpha}$ can be obtained by the embedding look-up operation:
\begin{equation}
\bm{e}_{\alpha} = \bm{E}_{\alpha} [\bm{x}_{\alpha}],
\end{equation}
where $\bm{E}_{\alpha} \in R^{v_\alpha\times d}$ is the embedding matrix for the $\alpha$th feature, the vocabulary size is $v_\alpha$ and embedding size is $d$. $\bm{E}_{\alpha}[\cdot]$ denotes look-up operation. For example, minute of week is embedded with the vocabulary size equals to 10080 and embedding size 8.

\textbf{Continuous features}: For the numerical fields, we transform numerical features to categorical features first. The embedding look-up for a numerical feature $\bm{x}_\beta$ can be written as 
\begin{equation}
\bm{e}_\beta = \bm{E}_\beta[Q(\bm{x}_\beta)],
\end{equation}
where $Q(\cdot)$ is the quantile bucketizing function and $\bm{E}_\beta \in R^{v_\beta\times d}$ is the embedding matrix with $v_\beta$ buckets after discretization. For example speed is bucketized into 256 quantiles, which led to better accuracy than using them directly. We found that using quantile buckets provided better accuracy than equal-width buckets, similar to other literatures have suggested \cite{saberian2019gradient}. One explanation is that for any fixed number of buckets, quantile buckets convey the most information in bits about the original feature value compared to any other bucketing scheme.

\textbf{Geospatial features}: We transform geospatial features, like longitudes and latitudes, using geohashing and multiple feature hashing. The idea is to first obtain a unique string to represent the 2D geospatial information and then map the string to a unique index for embedding look-ups. Therefore, the embedding for a pair of longitude and latitude $\bm{x}_k$ can be obtained by 
\begin{equation}
\bm{e}_k = \bm{E}_k[H(\bm{x}_k)],
\end{equation}
where $H(\cdot)$ is introduced in section \ref{subsec:hashing}.

\subsubsection{Geohash}
\label{subsec:geohash}
\begin{figure}[h]
  \centering
  \includegraphics[width=\linewidth]{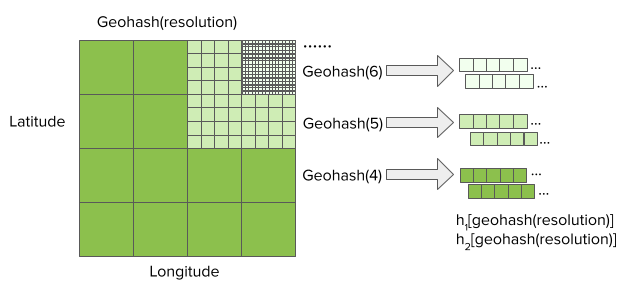}
  \caption{An illustration of multi-resolution geohashes with multiple feature hashing using independent hash functions $h_1$ and $h_2$.}
  \label{fig:location_grids}
\end{figure}

Geospatial longitudes and latitudes are key features for ETA predictions. However, they are distributed very unevenly over the globe and contain information at multiple spatial resolutions. We therefore use geohashes\cite{geohash} to map locations to multiple resolution grids based on latitudes and longitudes. As the resolution increases, the number of distinct grid cells grows exponentially and the average amount of data in each grid cell decreases proportionally. This is illustrated in Figure \ref{fig:location_grids}. 

We first use the geohash function $geohash(lat, lng, u)$ to obtain a length $u$ geohash string from a $(lat, lng)$ pair. The geohash function is described below:
\begin{itemize}
\item Map $lat$ and $lng$ into [0, 1] floats.
\item Scale the floats to $[0, 2^{32}]$ and cast to 32 bit integers.
\item Interleave the 32-bits from $lat$ and $lng$ into one 64 bit integer.
\item Base32 encode the 64-bit integer and truncate to a string in which each character represents $5u$ bits. 
\end{itemize}

\subsubsection{Feature hashing}
\label{subsec:hashing}
After we obtain the encoded geohash strings for each location and we must map the string to an index. We explored two different strategies for mapping these encoded geohash strings to indexes:

\begin{itemize}
    \item \textit{Exact indexing}. This strategy maps each grid cell to a dedicated embedding. This takes up the most space due to the exponential increase in cardinality with geohash precision. 
  \item \textit{Multiple feature hashing}. This strategy, described in Algorithm \ref{alg:hashall}, extends feature hashing by mapping each grid cell to multiple compact ranges of bins using independent hash functions, thus mitigating the effect of collisions when using only one hash. 
\end{itemize}

In the model, we include the geohash indexes of the begin and end location of a request separately and together as features. Taking a request that begins at $o=(lat_o, lng_o)$ and ends at $d=(lat_d, lng_d$) as an example, we use Algorithm \ref{alg:hashall} to obtain two independent hash buckets each for the origin $h_{o}$, destination $h_{d}$, and origin-destination pair $h_{od}$. We define independent hashing functions $h_1(x)$ and $h_2(x)$ as instances of MurmurHash3 \cite{MurmurHash3} with independent seeds. In addition, we use Algorithm \ref{alg:hashall} to create geospatial features at multiple resolutions $u \in \{4, 5, 6, 7\}$. The motivation is that using a granular geohash grid will provide more accurate location information but suffer from more severe sparsity issues. Using multiple resolutions can help alleviate the sparsity issue.   

\begin{algorithm}
\caption{Map geospatial features to indexes}\label{alg:hashall}
\begin{algorithmic}
\State Inputs: origin $o$, destination $d$, geohash resolution $u$
\State Outputs: independent hash bin pairs for origin  $h_{o}$, destination $h_{d}$, and origin-destination $h_{od}$
\State $H(x) \to (h_1(x), h_2(x))$
\State $H(geohash(o, u)) \to h_{o}$
\State $H(geohash(d, u)) \to h_{d}$
\State $H(geohash(o, u), geohash(d, u)) \to h_{od}$
\end{algorithmic}
\end{algorithm}

\subsection{Two-layer Module}
\label{section:encodermodule}
The DeeprETANet is a wide and shallow network with only two layers besides of the embedding layer. The first layer is a linear transformer layer and the second layer is a fully connected layer with calibration. The first linear transformer layer aims to learn the interaction of geospatial and temporal embeddings. The second calibration layer aims to adjust bias from various request types.

\subsubsection{Interaction layer}
In the travel time estimation literature \cite{gao2021adeep} and recommender system literature \cite{guo2017deepfm, wang2017deep, cheng2016wide}, many neural network models follow an embedding and feature interaction paradigm, in which a cross-product is used to learn the feature interaction. In DeeprETANet we learn the feature interactions via the linear self-attention \cite{katharopoulos2020transformers}, which is a sequence-to-sequence operation that takes in a sequence of vectors and produces a re-weighted sequence of vectors. In a language model, each vector represents a single word token, but in the case of travel time estimation, each vector represents a single feature. Unlike the sequence-to-sequence prediction, the order of the vector does not matter in our case, as the order of the input features does not make a difference. Therefore, we do not include the positional encoding in our design. 

We first tried the original self-attention implementation in the Transformer paper \cite{vaswani2017attention}. It uncovers pairwise interactions among the $L$ features by explicitly computing a $L*L$ attention matrix of pairwise dot products, using the softmax of these scaled dot-products to re-weight the features. When the self-attention layer processes each feature, it looks at every other feature in the input for clues and outputs the representation of this feature is a weighted sum of all features. Through this way, we can bake the understanding of all the temporal and spatial features into the one feature currently being processed. Taking trips from an origin A to a destination B as an example, the model inputs are vectors of the time, the location, the traffic condition and the distance between A to B. The linear transformer layer takes in the inputs and scales the importance of the distance given the time, the location and the traffic condition.

\begin{figure}[h]
  \centering
  \includegraphics[width=\linewidth]{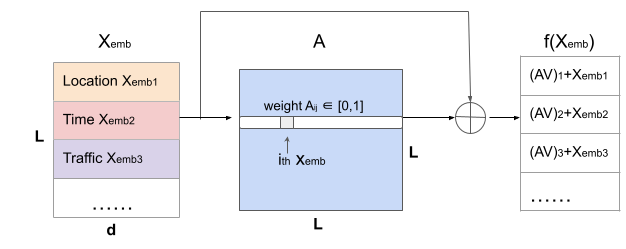}
  \caption{One interaction layer}
  \label{fig:atten_layer}
\end{figure}

For each ETA request, the feature embeddings are denoted as $X_{emb} \in R^{L*d}$ where $L$ is the number of feature embeddings and $d$ is the embedding dimension, $L \gg d$. The query, key and value in the self-attention is defined as  
\begin{align}
Q &=  X_{emb}W_q, \nonumber\\
K &=  X_{emb}W_k, \nonumber\\
V &=  X_{emb}W_v,
\end{align}
where $W_q, W_k, W_v \in R^{d*d}$. The attention is calculated as 
\begin{align}
A_{ij} &= \frac{\exp(QK^T/\sqrt{d} )_{i,j}}{\sum_{j=1}^L{\exp(QK^T/\sqrt{d} )_{i,j}}},
\end{align}
where $A\in R^{L*L}$. Then we use the attention matrix $A$ to calculate the output of the interaction layer: 
\begin{align}
f(X_{emb}) &= AV+X_{emb},
\label{eqn:interac}
\end{align}
using a residual structure. This interaction layer is also illustrated in Figure \ref{fig:atten_layer}. In the case that the embedding dimension is not equal to the dimension of value $V$, a linear layer can be used to transform the shape.

While the original transformer layer provided good accuracy, it was too slow to meet the latency requirements for online real-time serving. The original self-attention has quadratic time complexity, because it computes a $L \times L$ attention matrix. There have been multiple research works that linearize the self-attention calculation, for example linear transformer \cite{katharopoulos2020transformers}, linformer \cite{wang2020linformer}, performer \cite{choromanski2020rethinking}. After experimentation, we chose to use the linear transformer to speed up the computation. For the $i$th row of the weighted value matrix, 
\begin{align}
V_i^{\prime} &= \frac{\sum_{j=1}^L\phi(Q_i)^T\phi(K_j)V_j}{\sum_{j=1}^L\phi(Q_i)^T\phi(K_j)} \nonumber\\
&=\frac{\phi(Q_i)^T\sum_{j=1}^L\phi(K_j)V_j}{\phi(Q_i)^T\sum_{j=1}^L\phi(K_j)}
\end{align}
where the feature map $\phi(x) =elu(x)+1 = max(\alpha(e^x-1), 0) + 1$. Then the one linear iteration layer is
\begin{align}
f(X_{emb}) &= V^{\prime}+X_{emb},
\label{eqn:linearinterac}
\end{align}
It is evident that the computational cost of equation \ref{eqn:interac} is $O(L^2d)$ and the cost of equation \ref{eqn:linearinterac} is $O(Ld^2)$, assuming the feature map has the same dimension as the value. When $L\gg d$, which is a common case for travel time estimation, the linear transformer is much faster. 

\subsubsection{Calibration layer}
We implemented two ways to deal with data heterogeneity. One is to embed request types for learning the interaction between the type features and others via the interaction layer. The other way is through a calibration layer. The calibration layer consists of a fully connected layer and some bias parameters for each request type. Suppose $b_{j}$ denotes the bias of the $j$th ETA request type, $b_{j}$ is learned from a linear layer where the input is the one-hot encoded type features. Then, the DeeprETA residual of the $i$the request and $j$th type can be estimated as
\begin{equation}
 \hat{r}_{\mathrm{ij}} = \hat{f}_2(\hat{f}(X_{i_{\mathrm{emb}}})) + \hat{b}_j(X_{i_{\mathrm{type}}}), 
\end{equation}
where $f(\cdot)$ stands for the interaction layer and $f_2(\cdot)$ stands for a fully connected layer.

By implementing the calibration layer we can adjust the raw prediction by accounting for mean-shift differences across request types. This improves prediction accuracy with minimal latency increase. We also tried the multi-heads structure and mixture of expert structure \cite{shazeer2017outrageously}. However the latency does not meet the requirement. We also employ a few tricks to further improve prediction accuracy, such as using $ReLU(\cdot)=max(0, \cdot)$ at output to force predicted ETA to be positive and clamping to reduce the effect of extreme values.

\section{Model Training and Serving}
\label{section:train}
In this section, we introduce the training loss function and the serving details. The ETA downstream consumers require different metrics to evaluate the ETA predictions. That means optimizing for a single point estimate is not ideal. For example, when ETA is used for calculating fares, mean ETA error is important. While for evaluating delivery ETA requests, not only the mean absolute ETA error, but also the 95th quantile is important. Extreme ETA errors will result in bad user experiences. Therefore, to meet diverse business goals, DeeprETA uses a customized loss function, asymmetric Huber loss in equation \ref{eqn: asymhuber}, which is robust to outliers and can balance a range of commonly used point estimates metrics. 

\begin{align}
 \mathcal{L}(\delta,\Theta;(\bm{q},y_0),y) &=  
 \begin{cases}
 \frac{1}{2}(y-\hat{y})^2,& |y - \hat{y}| < \delta \\
 \delta |y - \hat{y}| -\frac{1}{2}  \delta ^2,& |y - \hat{y}| \geq \delta,
 \end{cases}
\label{eqn: huber}
\end{align}

\begin{align}
 \mathcal{L}(\omega,\delta,\Theta;(\bm{q},y_0),y) &=  
  \begin{cases}
\omega \mathcal{L}(\delta, \Theta;(\bm{q},y_0),y), & y < \hat{y} \\
(1-\omega) \mathcal{L}(\delta, \Theta;(\bm{q},y_0),y),&y \geq \hat{y},
 \end{cases}
\label{eqn: asymhuber}
\end{align}
where $\omega\in [0,1]$, $\delta>0$, $\Theta$ denotes the model parameters.

Unlike symmetric loss functions, an asymmetric loss function applies a different weight to underprediction vs overprediction, which is useful in ETA cases where being early is not as costly as being late. We started with a Huber loss \cite{huber1964robust} as a base loss function, since it is less sensitive to outliers, and extended it by making it asymmetric. The loss function has two parameters, $\delta$ and $\omega$, that control the degree of robustness to outliers and the degree of asymmetry respectively. By varying $\delta$, squared error and absolute error can be smoothly interpolated, with the latter being less sensitive to outliers. By varying $\omega$, we can control the relative cost of underprediction vs overprediction, which is useful in situations where being a minute late is worse than being a minute early in ETA predictions. These parameters not only make it possible to mimic other commonly used regression loss functions, but also make it possible to tailor the point estimate produced by the model to meet diverse business goals.     

We retrain and validate the model performance weekly using an auto retraining workflow. We employ the Nvidia Quadro RTX 5000 GPUs for training. Once trained, these models are deployed onto the serving platform. Production requests are served by a fleet of servers with 4 cores per host. 


\section{Experiments}
In this section, we evaluate the performance of DeeprETANet through offline experiments, online tests and embedding analysis. In the offline experiments, we compare DeeprETA with three other alternative methods. We also compare ablated variants of DeeprETA to validate the components of the model.

\subsection{Baseline methods}
To demonstrate the performance of the post-processing system, we compare several machine learning models with the routing engine ETA. Then to demonstrate the performance of the proposed DeeprETANet, we choose several frequently-used regression models as the baseline methods, specifically XGBoost\cite{chen2016xgboost}, ResNet\cite{he2016deep} and Hammock\cite{saberian2019gradient}. 

\textbf{Routing engine ETA} is a rule-based solution. It estimates the travel time along a given path as a weighted sum of the expected travel times along each segment. This method provides the fastest ETA inference, therefore it is widely used in map services and ride hailing platforms for route planning and recommendation. However, the objective of RE-ETA is find the best path, which is different from estimating an end-to-end ETA. The prediction accuracy is not the best due to re-routing and noisy data.

\textbf{XGBoost} is a decision-tree-based ensemble machine learning algorithm that uses a gradient boosting framework \cite{chen2016xgboost}. XGBoost is widely considered best-in-class for modeling tabular data. The parallelized implementation and efficient usage of hardware resources makes this method  appealing for applied machine learning. Before DeeprETA, the production ETA model at Uber was a group of XGBoosts with each trained on one mega-region . In this experiment, we use this group of XGBoosts and tuned its the hyperparameters: maximum depth to grow the tree is 15, maximum number of tree to grow is 250, the subsample ratio is set to 1 so that it will use all training data prior to growing trees. The learning rate is set to 0.1 and $L_2$ regularization is added on weights.

\textbf{ResNet} is a neural network model which introduces an identity shortcut connection that skips one or more layers. ResNet learns the residual mapping with reference to the layer inputs. In the experiment, the hyperparameters are tuned and we set the hidden sizes to be 512 and use 4 residual blocks. 

\textbf{HammockNet} is a method to build a neural network that is similar to an ensemble of decision trees \cite{saberian2019gradient}, i.e., equivalent to learning tree leafs internally or externally by specifying the feature discretization thresholds. We choose this method as a baseline since it has been shown to be competitive with XGBoost models on tabular data. We implemented the hammock model by using quantized and one-hot encoded inputs, and connected them by two fully connected layer with layer size [800, 100]. 

\textbf{DeeprETANet and its variants} DeeprETANet has two layers after the embedding layer. The linear transformer layer has the keys, queries and values dims to be 4. The fully connected layer has the layer size of 2048. We also evaluate two variants of DeeprETANet as an ablation study. One variant is without feature hashing, i.e. simply using the geohash function and indexing the geohash string. The other variant is without the calibration layer. For all the neural network models in our experiment, we use $ReLU(\cdot)$ as the activation function, Adam as the optimizer and relative cosine annealing learning rate scheduler.

\subsection{Metrics} 
We evaluate the accuracy using mean and quantile error metrics, including the mean absolute error (MAE), 50th percentile absolute error (p50 error) and 95th percentile absolute error (p95 error). We calculate the relative improvement by comparing the error of the proposed method to the error of the RE-ETA. For example the relative MAE improvement is calculated as  
\begin{equation}
    \frac{\frac{1}{n}\sum_{i=1}^n |y_i - \hat{y}_{0i}| - \frac{1}{n}\sum_{i=1}^n |y_i - \hat{y}| }{\frac{1}{n}\sum_{i=1}^n |y_i - \hat{y}_{0i}|}.
\label{eqn:eval}
\end{equation}
The relative p50 error and p95 error are calculated similarly. 

To evaluate the inference speed, we perform offline profiling and an online latency tests. As part of offline profiling, we use batch inference time as an approximation of the model speed to pre-screen models. In the online test, the latency, query per second (QPS) and CPU metrics are evaluated in the production environment. QPS indicates the throughput capacity and latency is the time between making a request and the completion of the response. 

\subsection{Dataset}
The dataset consists of global ETA requests from Uber's platform. The global data has two types of requests, one is ride-hailing and the other is eats delivery. We removed outliers with travel time greater than 2 hours. The collection time window is from September 13th, 2021 to October 1st, 2021 and the data are split to training, validation and test set sequentially. Training and validation sets have 14 days of data with a 90\%/10\% sequential split, and test set has 7 days of data. The training data has approximately 1.4 billion ETA requests. We train all models with the context features, temporal features, geospatial features, trip features and traffic features. The label is the ATA.

\subsection{Offline experiments}
In the offline experiments, all models are trained on the same training dataset and evaluated on the same test dataset. Training was on two Nvidia Quadro RTX 5000 GPUs. We implemented all neural network models in Python with the Pytorch framework \cite{NEURIPS2019_9015}. In this section, we first show the global results across different models, then analyze the city-level results on 4 major cities across the world.

Table \ref{tab:globaleval} shows the global evaluation results. The evaluation metrics are the relative error improvement from the RE-ETA error, shown in equation \ref{eqn:eval}. Specifically, we evaluate the relative MAE, relative p50 error and relative p95 error. DeeprETANet achieved the lowest MAE and p50 error. For the p95 error, DeeprETANet with and without feature hashing has similar performance. This result indicates that richer geospatial embeddings improve performance in typical cases but may not improve extreme errors. Comparing DeeprETANet with its two variants, removing the calibration and removing the feature hashing, we see both will degrade the model accuracy. 

\begin{table}
  \caption{Evaluation of ETA predictions}
  \label{tab:globaleval}
  \resizebox{\columnwidth}{!}{
  \begin{tabular}{cccc}
    \toprule
  Relative improv   & MAE improv(\%) & p50 improv(\%)& p95 improv(\%)\\
    \midrule
XGBoost &5.07&	1.79&	5.98\\
ResNet &7.39&	6.19&	7.27\\
HammockNet &7.58&	6.12&	7.56\\
\midrule
DeeprETANet & $\bm{7.83}$ &	$\bm{6.25}$ &	$\bm{7.99}$\\
$\dots$ w/o calibration  &7.73&	6.12&	7.85\\
$\dots$ w/o feature hashing  &7.41&	5.10& $\bm{7.99}$\\
\bottomrule
\end{tabular}}
\end{table}

ETAs of the delivery and rides requests have quite different distributions. In Table \ref{tab:globalevalseg}, we evaluate the DeeprETANet global performance on delivery and rides requests separately. Interestingly, rides requests have significantly lower p50 error and delivery requests have significantly lower p95 error.
\begin{table}
  \caption{DeeprETANet improvements on rides and delivery requests}
  \label{tab:globalevalseg}
  \begin{tabular}{cccc}
    \toprule
& MAE improv(\%) & p50 improv(\%)& p95 improv(\%)
\\
    \midrule
Delivery &7.46&	2.26&	11.05\\
Rides &7.86&	6.25&	7.62\\
  \bottomrule
\end{tabular}
\end{table}

Table \ref{tab:complexity} lists the number of parameters in three deep neural networks: ResNet, HammockNet and DeeprETANet. The three models all have around 140 million parameters and more than 99\% of the parameters lie in the embedding layers.

\begin{table}
  \caption{Complexity analysis of Neural Network Model }
  \label{tab:complexity}
  \begin{tabular}{ccc}
    \toprule
     $\#$parameters & total ($\times 10^6$) & non-embedding ($\times 10^6$)
\\
    \midrule
ResNet &141.6&1.2 \\
HammockNet &140.8&0.5 \\
DeeprETANet &141.2&0.5 \\
  \bottomrule
\end{tabular}
\end{table}

\begin{figure}[h]
     \centering
     \begin{subfigure}[b]{0.475\linewidth}
         \centering
         \includegraphics[width=\linewidth]{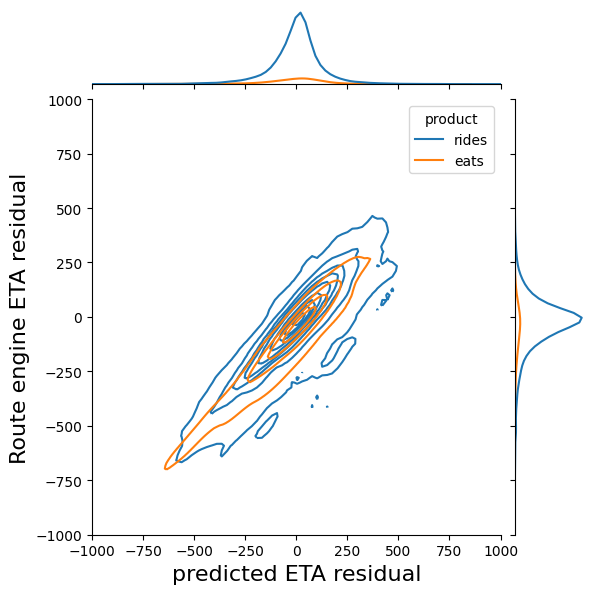}
         \caption{New York City}
     \end{subfigure}
     \hfill
     \begin{subfigure}[b]{0.475\linewidth}
         \centering
         \includegraphics[width=\linewidth]{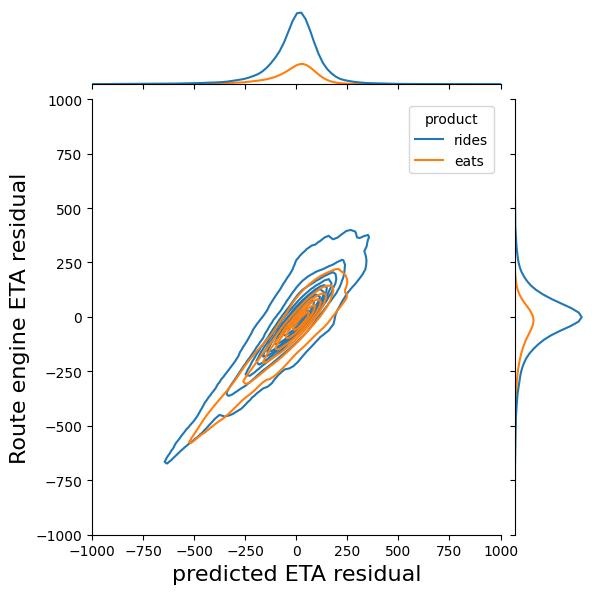}
         \caption{Los Angeles}
     \end{subfigure}
\vskip\baselineskip
     \begin{subfigure}[b]{0.475\linewidth}
         \centering
         \includegraphics[width=\linewidth]{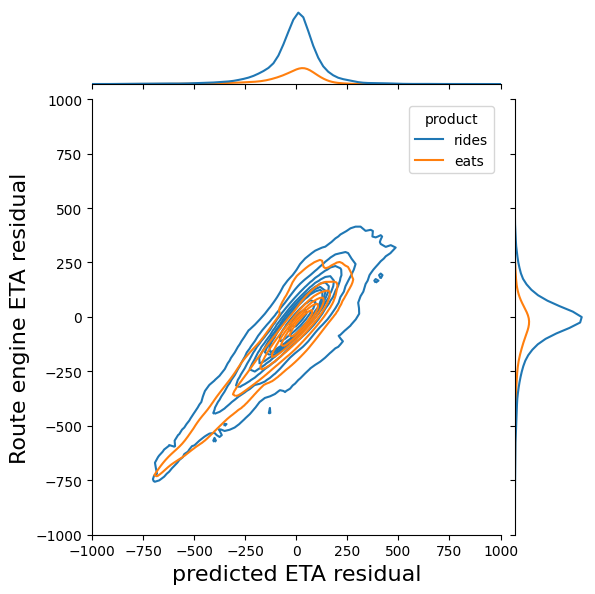}
         \caption{London}
     \end{subfigure}
    \begin{subfigure}[b]{0.475\linewidth}
         \centering
         \includegraphics[width=\linewidth]{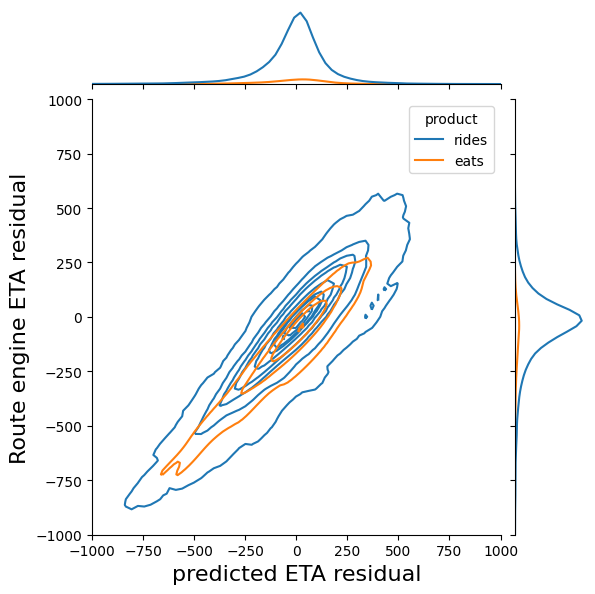}
         \caption{Mexico City}
     \end{subfigure}
     \hfill
        \caption{City level bivariate graphs of 1\% sampled RE-ETA residuals and predicted DeeprETA residuals}
        \label{fig:4residuals}
\end{figure}

In addition, we analyze the experiment results on 4 major cities globally. All the models are trained and evaluated using the global data. We share the city-level data details and model evaluation as a case study to better understand the regional performance. Table \ref{tab:citystats} shows the number of delivery and rides ETA requests at each city. Table \ref{tab:cityeval} shows the relative MAE improvement from the 4 models in each city. Overall New York City improves the most from using DeeprETANet among the 4 cities. We can also see the difference before and after DeeprETA post-processing from Figure \ref{fig:4residuals}, which visualize the bivariate distribution of the RE-ETA residual and the predicted ETA residual on a 1\% sampled data. Although rides and delivery requests have quite different residual distributions, after post-processing the mean residual is closer to 0.

\begin{table}
  \caption{City level ETA request in the training data}
  \label{tab:citystats}
     \resizebox{\columnwidth}{!}{%
  \begin{tabular}{ccccc}
    \toprule
   Count ($\times 10^6$) &  New York City & Los Angeles & London & Mexico City\\
    \midrule
Delivery &19.94&17.46&18.33&16.11\\
Rides &138.62&58.32&102.84&180.39\\
  \bottomrule
\end{tabular}}
\end{table}

\begin{table}
  \caption{City level evaluation of ETA predictions}
  \label{tab:cityeval}
    \resizebox{\columnwidth}{!}{%
\begin{tabular}{ccccc}
    \toprule
  Relative MAE improv(\%) &  New York City & Los Angeles & London & Mexico City
\\
    \midrule
XGBoost &4.09&	4.84&	1.39&	3.93\\
ResNet &7.02&	6.29&	4.09&	5.86\\
HammockNet &7.13&	6.41&	4.28&	6.31\\
DeeprETANet &$\bm{7.20}$&	$\bm{6.80}$&	$\bm{4.80}$& 	$\bm{6.60}$\\
  \bottomrule
\end{tabular}%
}
\end{table}

\subsection{Online experiments}
\label{subsec:online}
Online shadow experiments were conducted globally to verify the performance of DeeprETA over a 2-week window. The launch criteria is to improve ETA accuracy from an online XGBoost model. The improvement is measured by MAE, p50 error and p95 error across mega-regions globally. From the online results in Table \ref{tab:onlineevalseg}, DeeprETA outperforms the in production XGBoost on both rides and delivery requests. This result is aggregated from multiple launches of DeeprETA. 
\begin{table}
  \caption{DeeprETANet improvements over XBGoost on rides and delivery requests in online tests}
  \label{tab:onlineevalseg}
  \begin{tabular}{cccc}
    \toprule
& MAE improv(\%) & p50 improv(\%)& p95 improv(\%)
\\
    \midrule
Delivery &2.91&4.74&	3.92\\
Rides &2.66&	4.15&3.23 \\
  \bottomrule
\end{tabular}
\end{table}

\begin{figure}[h]
    \centering
    \includegraphics[width=\linewidth]{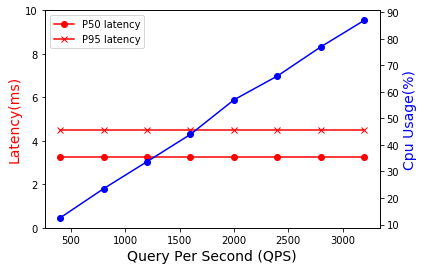}
    \caption{Latency, QPS and CPU usage}
    \label{fig:latency}
\end{figure}

To be able to meet production serving requirements, DeeprETA needs to be fast so as not to add too much latency to an ETA request. This post-processing system has the highest QPS at Uber. Figure \ref{fig:latency} visualizes the average latency under different loads. The result shows that DeeprETA is stable when handling different QPS loads. The median latency is 3.25ms and the 95th percentile is 4ms. 

\subsection{Embedding Analysis}
We select 14 geospatial embeddings and 1 temporal embedding, then project them separately onto a two dimensional space using t-SNE \cite{van2008visualizing}. The visualization is based on randomly sampled data points of rides and delivery requests in San Francisco. 

\begin{figure}[h]
     \centering
     \begin{subfigure}[b]{0.475\linewidth}
         \centering
         \includegraphics[width=\linewidth]{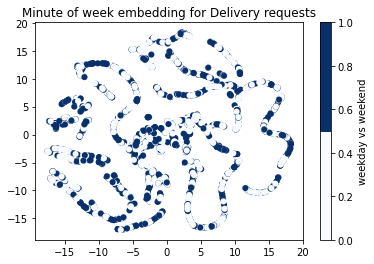}
     \end{subfigure}
     \hfill
     \begin{subfigure}[b]{0.475\linewidth}
         \centering
         \includegraphics[width=\linewidth]{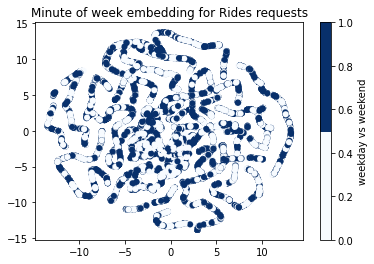}
     \end{subfigure}
    \caption{T-SNE time embedding visualization}
    \label{fig:timeemb}
\end{figure}

\begin{figure}[h]
     \centering
     \begin{subfigure}[b]{0.475\linewidth}
         \centering
         \includegraphics[width=\linewidth]{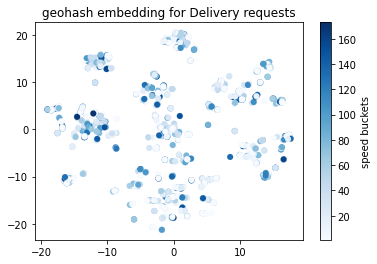}
     \end{subfigure}
     \hfill
     \begin{subfigure}[b]{0.475\linewidth}
         \centering
         \includegraphics[width=\linewidth]{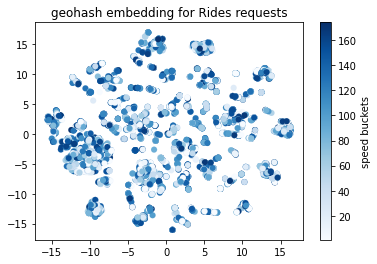}
     \end{subfigure}
    \caption{T-SNE geohash embedding visualization}
    \label{fig:geoemb}
\end{figure}

Figure \ref{fig:timeemb} shows the minute-of-week embedding. The color indicates whether it is a weekday or a weekend. We can observe that the minute-of-week embedding has local continuity and is a one-dimensional manifold. However there is no clear patterns of the weekdays vs weekends effect. Figure \ref{fig:geoemb} shows the visualization of the geohash embeddings. The color indicates the speed buckets. We can observe that the geohash embeddings are locally clustered based on the geo location. Similar locations are represented by similar positions in two-dimensional space. Interestingly, the high speed locations of rides and delivery requests do not all overlap.

\section{Conclusion}
In this paper we introduced the DeeprETA post-processing system, which provides an accurate and fast travel time prediction. It is launched in production and now serves all 4-wheel ETA prediction requests at Uber. We presented offline and online evaluations which all show significant quantitative improvements over traditional machine learning models. We provide our understanding and prescriptive advice on modeling geospatial-temporal data. We believe our findings will be beneficial to researchers and industrial practitioners who apply deep neural networks to similar geospatial-temporal problems. One limitation of our work is that we only evaluate the performance of our hybrid approach using Uber's proprietary routing engine and not with other 3rd party or open-source routing engines; that said, one of the benefits of our hybrid approach is that it is decoupled from the details of any particular routing engine implementation, and we expect that teams using other routing engines will be able to achieve similar accuracy improvements using our method. In the future, more work is underway to further expand the accuracy improvements by looking at every aspect of the modeling process including model architectures, loss functions and infrastructure improvements.


\bibliographystyle{ACM-Reference-Format}
\bibliography{sample-sigconf}


\appendix

\end{document}